\documentclass[review]{elsarticle}

\usepackage{lineno,hyperref}
\modulolinenumbers[5]

\journal{a journal}

\usepackage{graphicx}
\usepackage{amsmath,amssymb,amsfonts}
\usepackage{rotating}
\usepackage{listings}
\usepackage{bm}
\usepackage{multirow}
\usepackage[color]{changebar}
\usepackage{caption}
\usepackage{url}
\usepackage{natbib}
\usepackage{booktabs}
\usepackage{epstopdf}
\usepackage{subfigure}

\begin{document}

\begin{frontmatter}

\title{Explaining the Attention Mechanism of End-to-End Speech Recognition Using Decision Trees}



\author[addr3]{Yuanchao Wang}
\ead[url]{w907259849@bjfu.edu.cn}

\author[addr4]{Wenji Du}
\ead[url]{duwenji21@mails.ucas.ac.cn}

\author[addr3]{Chenghao Cai}
\ead[url]{chenghao.cai@outlook.com}

\author[addr3]{Yanyan Xu*}
\ead[url]{xuyanyan@bjfu.edu.cn}
\cortext[corr_author]{Corresponding author}

\address[addr3]{School of Information Science and Technology, Beijing Forestry University, Beijing, China}

\address[addr4]{Computer Network Information Center, Chinese Academy of Sciences, Beijing, China}


\begin{abstract}
The attention mechanism has largely improved the performance of end-to-end speech recognition systems. However, the underlying behaviours of attention is not yet clearer. In this study, we use decision trees to explain how the attention mechanism impact itself in speech recognition. The results indicate that attention levels are largely impacted by their previous states rather than the encoder and decoder patterns. Additionally, the default attention mechanism seems to put more weights on closer states, but behaves poorly on modelling long-term dependencies of attention states.
\end{abstract}

\begin{keyword}
Attention Mechanism \sep Decision Trees \sep Automatic Speech Recognition \sep End-to-End Model
\end{keyword}

\end{frontmatter}


\section{Introduction}

Automatic Speech Recognition (ASR) \cite{DBLP:conf/icassp/ChanJLV16,DBLP:conf/icassp/WangMLLXMHTZZFZ20} means the use of algorithms to convert voices to texts. Traditional ASR systems usually consists of individual components such as acoustic models, lexicons and language models. As these components are constructed independently, additional effort is required to develop algorithms and collect data for each of the components. To solve this problem, end-to-end ASR systems, which are based on sequence-to-sequence models \cite{DBLP:conf/nips/SutskeverVL14,DBLP:conf/icml/AmodeiABCCCCCCD16,DBLP:conf/icassp/ChiuSWPNCKWRGJL18}, have been developed to convert acoustic data to texts directly. While the traditional ASR systems based on Gaussian Mixture Models and Hidden Markov Models (GMM-HMM) requires context-dependent signal pre-processing and force alignment to obtain input data and labels for supervised learning, the end-to-end ASR systems can directly use acoustic data and texts to perform supervised learning.

The attention mechanism \cite{DBLP:conf/nips/ChorowskiBSCB15,DBLP:conf/nips/VaswaniSPUJGKP17} is one of the essential functions to boost the performance of the end-to-end ASR systems. The attention mechanism can increase the impact of useful information and decrease the impact of useless information by applying different weights to specific districts of data. It is still unclear how the attention mechanism impacts final decisions of ASR. According to two past studies \cite{DBLP:conf/acl/SerranoS19,DBLP:conf/naacl/JainW19}, the attention mechanism may have two types of behaviours, which depend on the attention mechanism itself and a whole model, respectively. As the impact on attention mechanism itself may have more model-independent rules, in this article, we study how the attention mechanism of ASR impact itself. To explain the attention mechanism, we build an ASR system and use the Silas decision tree tool \cite{DBLP:journals/eswa/BrideCDDHMS21} to learn the distributions of attention weights and extract the relationships among the attention weights.

\section{Methods}

We train an ASR model based on the encoder-decoder architecture with the attention mechanism \cite{DBLP:conf/nips/ChorowskiBSCB15}. The encoder consists of two LSTM-RNNs. The decoder is a single LSTM-RNN. The attention mechanism is based on the hybrid structure. The whole encoder-decoder model is trained on the TIMIT training dataset. During each epoch of training, 200 speech files in the TIMIT dataset are randomly selected to train the whole encoder-decoder model. The training process terminates after 1,000 epochs.

In order to generate data for attention mechanism analysis, the TIMIT evaluation set is fed into the trained ASR model. We select 770 audio files that obtain the best phoneme error rate and extract their attention weight matrices in the ASR model. The extracted attention weight matrices are analysed by the following steps.
\begin{itemize}
\item[1] All attention weights in the attention weight matrices are sorted in ascending order. The sorted attention weights are averagely split into 10 domains. The domains are annotated with 10 levels that represent the strength of attention.
\item[2] Attention level matrices are produced by converting all of the attention weights in the attention weight matrices to their corresponding levels. As the size of an attention matrix subjects to the encoder output and the decoder output, the size of a attention level matrix is defined by the maximal size, i.e., $ 100 \times 659 $, where 100 is the maximal size of the encoder output, and 659 is the maximal size of the decoder output. All vacancies in the attention level matrices are filled by 0.
\item[3] To observe how the $i$th row of the attention level matrix is influenced by the $ (i-p),\ldots,i-1) $th rows, where $ i = (1, \ldots, 100) $ and $ p = (1, \ldots, 8) $, we produce a feature by concatenating the $ (i-p),\ldots,i-1) $th rows.
\item[4] Each attention level in the $ i $th row is converted to a label. An attention level higher than 5 is considered as “high”, while an attention level not higher than 5 is considered as “low”. The labels and the features together form a binary classification dataset.
\item[5] The dataset is shuffled and split into a training set and an evaluation set that consists of 80\% and 20\% of data, respectively. The training set is used to train 100 decision trees using the Silas tool. Each decision tree has a maximum depth of 64, and each leaf node has at least 64 training examples.
\item[6] The trained decision trees are scored on the evaluation set. We observe the scores, i.e., prediction accuracy, for each encoder state. Besides, we collected the decision conditions and their influence scores computed by Silas \cite{DBLP:journals/eswa/BrideCDDHMS21}.
\end{itemize}

\section{Results}

\begin{figure}
\centering
\begin{minipage}[t]{0.45\textwidth}
\centering
\includegraphics[width=6cm]{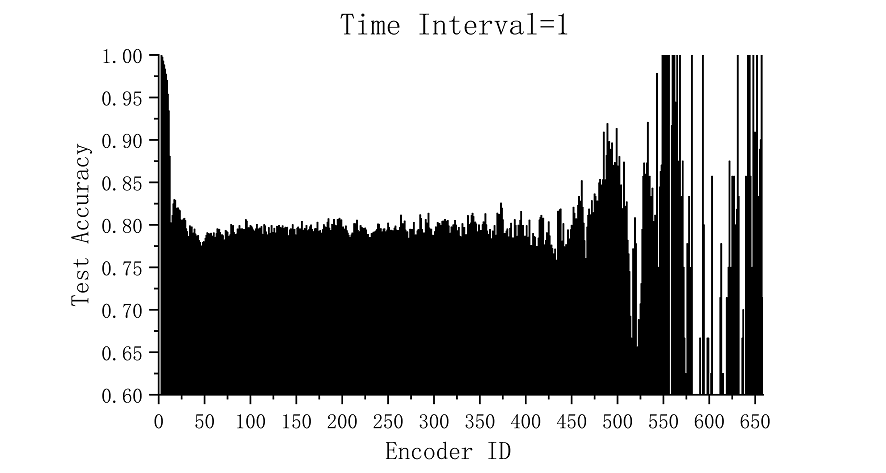}
\end{minipage}
~~
\begin{minipage}[t]{0.45\textwidth}
\centering
\includegraphics[width=6cm]{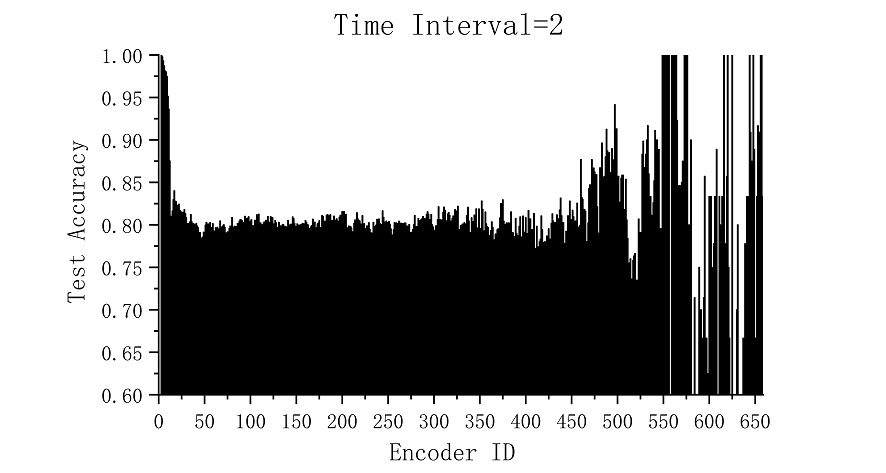}
\end{minipage}
\\
\begin{minipage}[t]{0.45\textwidth}
\centering
\includegraphics[width=6cm]{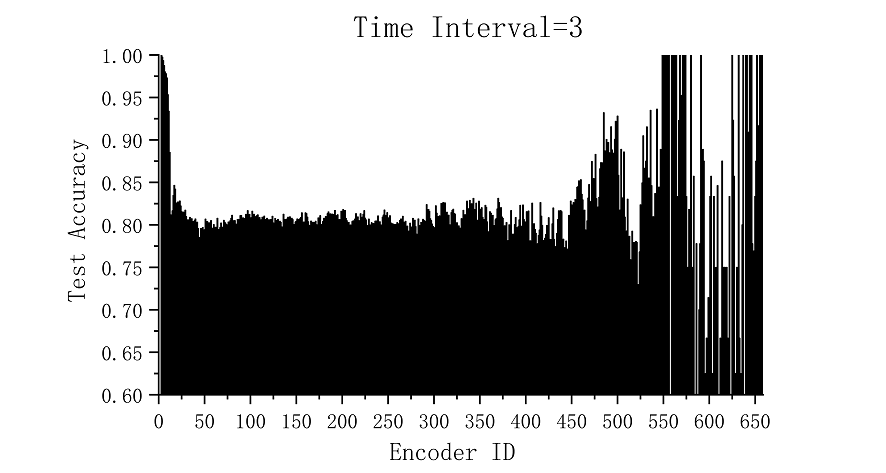}
\end{minipage}
~~
\begin{minipage}[t]{0.45\textwidth}
\centering
\includegraphics[width=6cm]{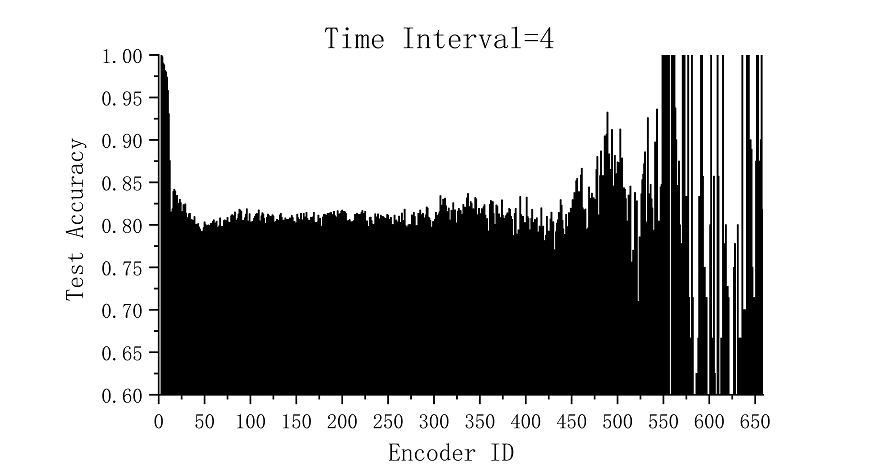}
\end{minipage}
\\
\begin{minipage}[t]{0.45\textwidth}
\centering
\includegraphics[width=6cm]{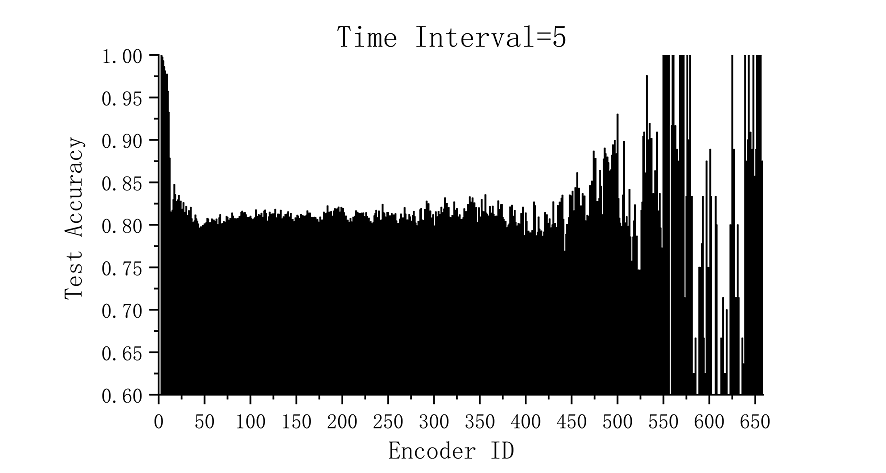}
\end{minipage}
~~
\begin{minipage}[t]{0.45\textwidth}
\centering
\includegraphics[width=6cm]{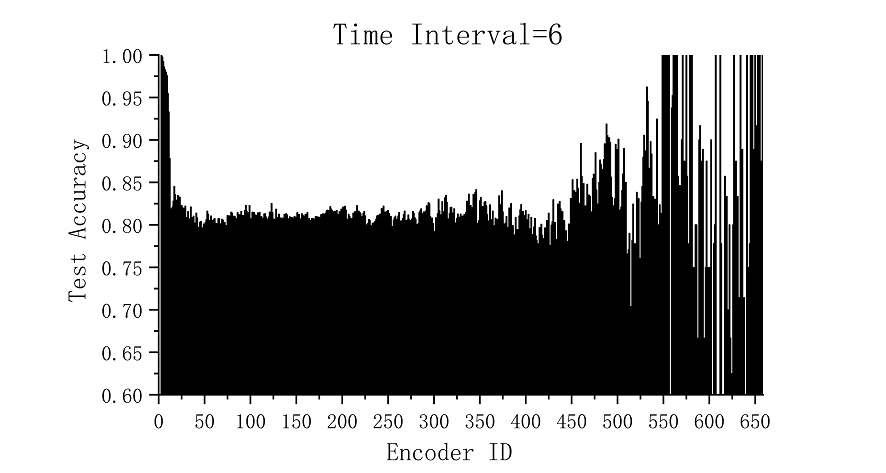}
\end{minipage}
\\
\begin{minipage}[t]{0.45\textwidth}
\centering
\includegraphics[width=6cm]{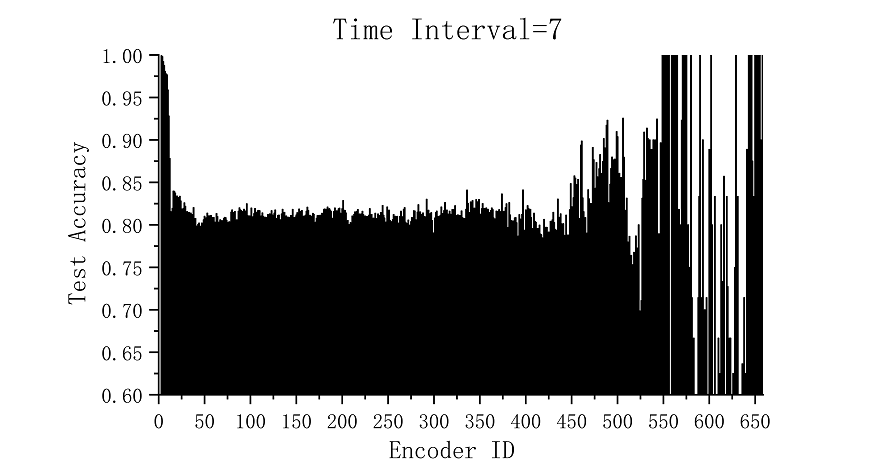}
\end{minipage}
~~
\begin{minipage}[t]{0.45\textwidth}
\centering
\includegraphics[width=6cm]{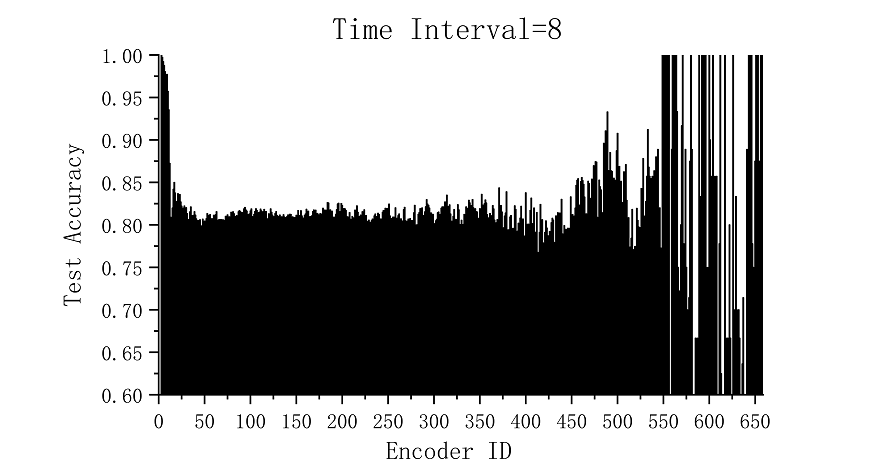}
\end{minipage}
\\
\caption{The Accuracy of Attention Level Prediction.}
\label{fig:attention-level-predition}
\end{figure}

\begin{figure}
\centering
\includegraphics[width=8cm]{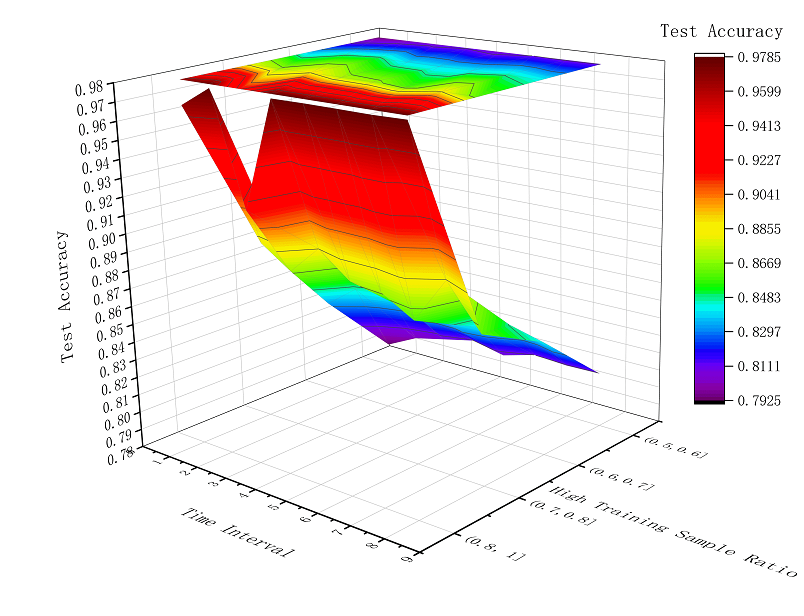}
\caption{The Data Distribution of Attention Levels.}
\label{fig:attention-level-data-distribution}
\end{figure}

Figure \ref{fig:attention-level-predition} shows the accuracy of attention level prediction. It is observable that the accuracy is high for small encoder IDs. This is probably because the smaller encoder IDs correspond to the beginnings of audio files that are almost silence. As the silence does not contain useful information, the attention on the silence is almost stable, which means that the attention level is relative level to predict. For most encoder IDs, the accuracy is around 80\%, which means that the attention is mostly predictable. For larger encoder IDs, the accuracy is unstable because of the lack of training data, i.e., most audio files do not use such a large number of encoder IDs. As a supplementary, Figure \ref{fig:attention-level-data-distribution} shows the data distribution of attention levels on all of the training data. It indicates that high level attention weights have positive impacts on the accuracy.

\begin{figure}
\centering
\includegraphics[width=8cm]{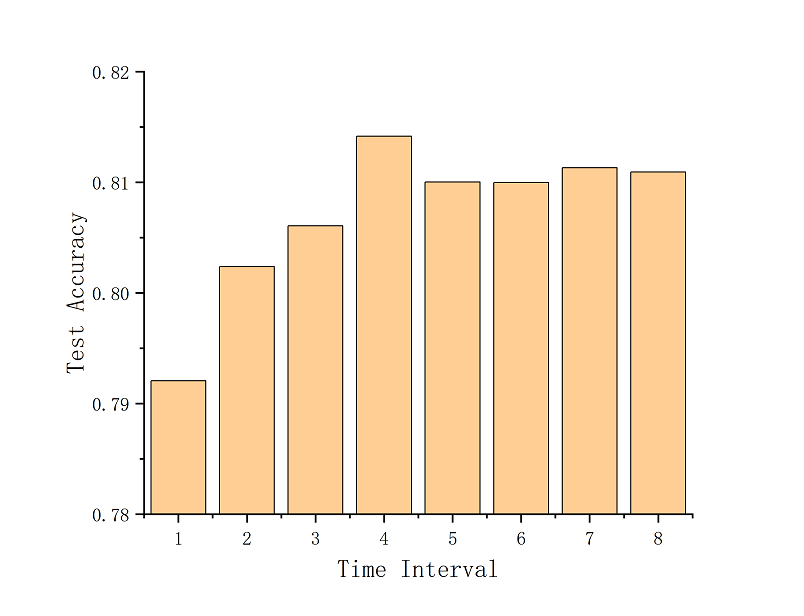}
\caption{The Accuracy of Attention Level Prediction with Respect to the Number of Previous States.}
\label{fig:attention-level-predition-prev-states}
\end{figure}

Figure \ref{fig:attention-level-predition-prev-states} shows the accuracy with respect to the number of previous states. It indicates that the accuracy is increasing as the number of precious states increases. Moreover, the previous four states have the highest impact on the accuracy. When the number of previous states is greater than four, more previous states cannot increase the accuracy.

\begin{figure}
\centering
\includegraphics[width=8cm]{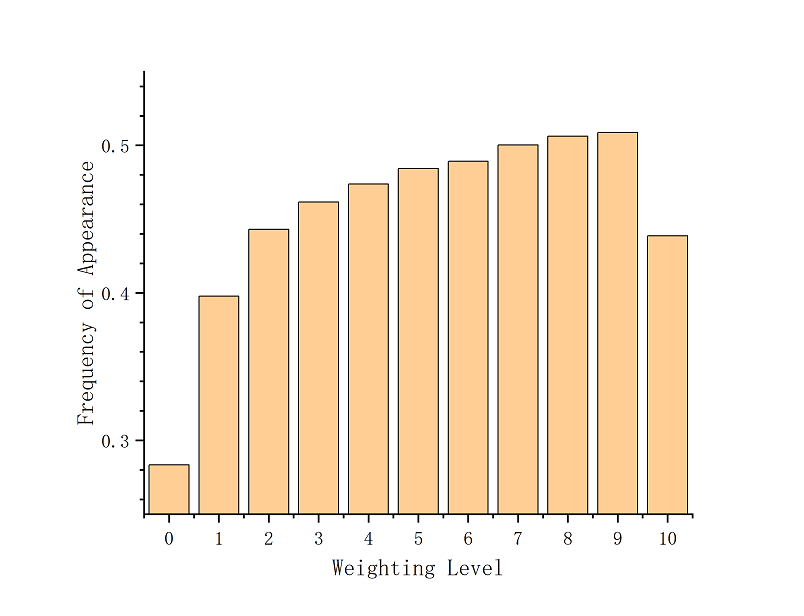}
\caption{The Frequencies of Attention Weight Levels on Decision Conditions.}
\label{fig:attention-level-frequency}
\end{figure}

Figure \ref{fig:attention-level-frequency} shows the frequencies of attention levels on decision conditions. It indicates that higher attention levels contribute more decision conditions, i.e., higher attention weights have larger impact on the future attention states.

\begin{figure}
\centering
\includegraphics[width=8cm]{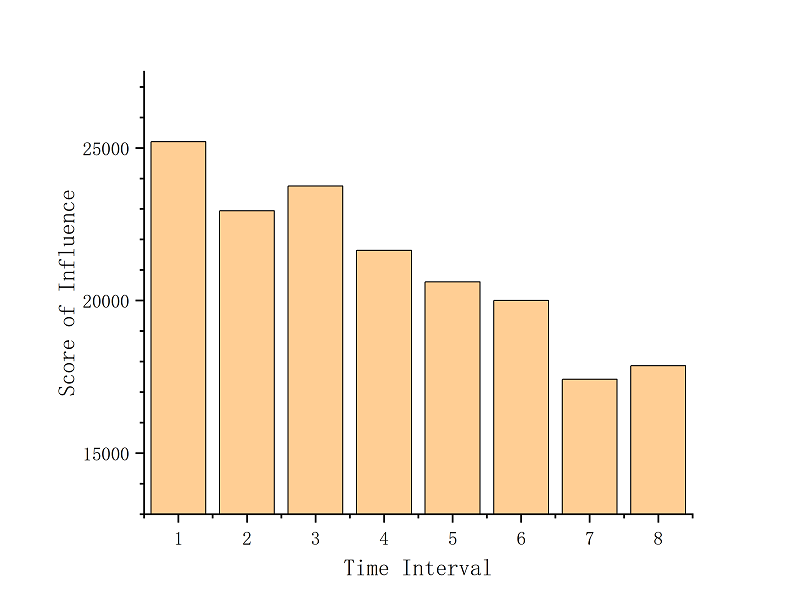}
\caption{The Average Influence Scores of Previous States.}
\label{fig:high-frequency-impact}
\end{figure}

Figure \ref{fig:high-frequency-impact} shows the average influence scores of previous states. It indicates a trend that the influence scores decrease when the time interval increases, which means that the nearer previous attention states have more impact on the current attention state. This phenomenon agrees with the results in Figure \ref{fig:attention-level-predition-prev-states}.

\section{Conclusion}

In this study, we have used decision trees to explain how the attention mechanism impact itself in end-to-end ASR models. The results show that the current attention state is mainly impacted by its previous attention states rather than the encoder and decoder states. It is possible that the attention mechanism on sequential tasks, e.g., speech recognition, is continuously impacted by its historical attention states. Moreover, the past four previous attention states have the highest impact on the current attention state. The influence scores keep decreasing when the time interval increases. However, in real ASR applications, time intervals are usually very large. The abovementioned phenomenon indicates that the attention mechanism should be improved by strengthening the attention on larger time intervals. This indicates a possible way to improve the attention mechanism in the future.

\bibliography{mybibfile}

\end{document}